%% file: MAIN_arxiv.tex
\documentclass[10pt,twocolumn,letterpaper]{article}

\usepackage{cvpr}              

\usepackage{graphicx}
\usepackage{amsmath}
\usepackage{amssymb}
\usepackage{booktabs}
\usepackage{paralist}
\usepackage{multirow}
\usepackage{color}
\usepackage{adjustbox}
\usepackage[accsupp]{axessibility} 
\usepackage[pagebackref,breaklinks,colorlinks]{hyperref}

\usepackage[capitalize]{cleveref}
\crefname{section}{Sec.}{Secs.}
\Crefname{section}{Section}{Sections}
\Crefname{table}{Table}{Tables}
\crefname{table}{Tab.}{Tabs.}
\newcommand*{\affaddr}[1]{#1}
\newcommand*{\affmark}[1][*]{\textsuperscript{#1}}
\newcommand*{\email}[1]{\texttt{#1}}

\begin{document}

\title{Efficient Remote Photoplethysmography with \\  Temporal Derivative Modules and Time-Shift Invariant Loss } 
\author{
Joaquim Comas\affmark[1], Adrià Ruiz\affmark[2], Federico Sukno\affmark[1]\\
\affaddr{\affmark[1] Department of Information and Communication Technologies,\\ Pompeu Fabra University, Barcelona, Spain}\\
\affaddr{\affmark[2] Seedtag, Madrid, Spain}\\
\small \email{ \affmark[1]  \{joaquim.comas,federico.sukno\}@upf.edu}\\
\small \email{\affmark[2] adriaruiz@seedtag.com}
} 

\maketitle
\input{Abstract/Abstract}
\input{Introduction/Introduction}
\input{Related_work/Related_work}
\input{Methodology/Methodology}

\input{Experiments/Experiments}

\input{Conclusions/Conclusions}
\input{Acknowledgment/Acknowledgment}

{\small
\bibliographystyle{ieee_fullname}
\bibliography{MAIN_arxiv}
}
\end{document}

%% file: Abstract/Abstract.tex
\begin{abstract}
   We present a lightweight neural model for remote heart rate estimation focused on the efficient spatio-temporal learning of facial photoplethysmography (PPG) based on $i)$ modelling of PPG dynamics by combinations of multiple convolutional derivatives, and $ii)$ increased flexibility of the model to learn possible offsets between the facial video PPG and the ground truth. PPG dynamics are modelled by a Temporal Derivative Module (TDM) constructed by the incremental aggregation of multiple convolutional derivatives, emulating a Taylor series expansion up to the desired order. Robustness to ground truth offsets is handled by the introduction of TALOS (\textbf{T}emporal \textbf{A}daptive \textbf{LO}cation \textbf{S}hift), a new temporal loss to train learning-based models. We verify the effectiveness of our model by reporting accuracy and efficiency metrics on the public PURE and UBFC-rPPG datasets. Compared to existing models, our approach shows competitive heart rate estimation accuracy with a much lower number of parameters and lower computational cost.

\end{abstract}

%% file: Introduction/Introduction.tex
\section{Introduction}
\label{sec:intro}

Camera-based measurement of human physiological signals has gained considerable interest in the research community in recent years. The monitoring of vital signals such as heart rate (HR), heart rate variability (HRV), respiration rate (RR), oxygen saturation (SpO2) or blood volume pulse (BVP) is crucial in the assessment of the physical and mental human state, and its potential applications \cite{ronca2021video, benezeth2018remote, liu20163d}. 
Traditionally, physiological signals are acquired using continuous bio-signal monitoring devices, which typically require physical contact between the sensor and the skin. However, such physical contact limits the mobility of the subjects and may also be intrusive to users in many scenarios, causing some biases during the acquisition stage. Different approaches have emerged to estimate vital signals without physical contact to minimize these limitations, among which, video-based methods, especially remote photoplethysmography (rPPG), have attracted much attention due to their low cost, non-invasive nature, and wide applicability \cite{poh2010advancements}. 

The rPPG or imaging photoplethysmography (iPPG) is the non-contact measurement of the PPG signal based on videos. The first rPPG measurements were proposed by Takano et al. \cite{takano2007heart} and Verkruysse et al. \cite{verkruysse2008remote} in 2007 and 2008, respectively, who demonstrated the feasibility of extracting the pulse signal using a conventional camera. Since then, several handcrafted methods have been proposed \cite{de2013robust,wang2015novel, wang2016algorithmic} mainly based on optical models of the skin. These approaches are usually sensitive to non-controlled scenarios, including head motion and illumination changes, and require a multi-stage process where some of the steps are difficult to adjust. 

The fast growth of deep learning (DL) techniques has led to more accurate and robust methods that have shown impressive results outperforming traditional approaches \cite{yu2019remote, perepelkina2020hearttrack, tsou2020siamese}.
A key aspect for such success relates to the enhanced ability of these approaches to adequately model the spatial and, more importantly, temporal information (dynamics) present in the facial videos. Indeed, the most successful approaches to date are based on the use of 3D Convolutional Neural Networks (3DCNNs) that offer great flexibility to model any interaction between spatial and temporal patterns at once, though at the expense of considerably increasing the number of trainable parameters and thus the complexity and size of the model. 

Alternatives to 3DCNNs have also been explored by combining spatial blocks based on 2DCNNs with additional modules or pre-processing operations that also take into account the temporal information of the signal. Notable examples include the use of normalized frame differences at the system's input \cite{chen2018deepphys} or after a number of convolutional layers \cite{yu2020autohr}, as well as Temporal Shift Modules (TSM) \cite{liu2020multi} or combinations of 2D and 1D convolutions \cite{liu2020general}. 

Even though the above alternatives avoid the use of the expensive 3D convolutions, achieving competitive performance in this way has proved challenging, requiring the clever combination of multiple modules to compensate for the noisy and unstable nature of rPPG dynamics, and resulting again in arguably large models. In this paper, we show that it is possible to model rPPG dynamics by means of local derivative filters, provided that they are applied to the appropriate spatial scale and to a sufficient order. This leads to a comparatively simpler architecture in which rPPG dynamics are captured by a chain of temporal derivative modules equipped with skip connections that emulates a Taylor series expansion up to a pre-established order. As shown in our experiments, our system is comparable to state-of-the-art approaches in terms of performance, but it is computationally more efficient and requires one or even two orders of magnitude less parameters.

In order to effectively exploit temporal information, another important aspect in deep learning approaches is the reliability of the ground-truth signals. While traditional techniques do not depend on the acquisition of the ground truth PPG because they directly estimate the rPPG signal from the skin pixels, data-driven methods need the ground-truth signal (e.g. from a pulsimeter) as a reference and assume that it is well synchronized with respect to the facial video. Nevertheless, these two modalities are often not correctly aligned, and there is an offset between the facial pulse signal and the ground truth signal, commonly caused by the Pulse Transit Time (PTT), whose exact value is unknown. A few recent works have highlighted the importance of such offset in performance, either by demonstrating improved results when the offset can be estimated beforehand and corrected \cite{zhan2020analysis} or by training models with modified loss functions that are more robust to the amount of ground-truth offset, although they do not model or estimate its value \cite{gideon2021way}.
In contrast, we propose a more intuitive temporal-invariant loss by modifying the standard mean square error (MSE) loss to account for the temporal offset of the ground truth and automatically estimate its value for each training sequence.

\subsection{Contribution}

In this work, we present a lightweight architecture to address both the efficient modelling of rPPG dynamics and the impact of temporal desynchronization between video and ground-truth physiological data. Firstly, we introduce a novel architecture in which rPPG dynamics are captured by a chain of temporal derivative modules equipped with skip connections that emulates a Taylor series expansion up to a pre-established order. Secondly, we present a new temporal loss, which we denote TALOS (\textbf{T}emporal \textbf{A}daptive \textbf{LO}cation \textbf{S}hift), that allows training of deep learning methods invariantly to temporal offsets of the ground-truth signal. Compared to existing models, our approach shows competitive heart rate estimation accuracy with a much lower number of parameters and lower computational cost.



%% file: Related_work/Related_work.tex
\section{Related work}
\label{sec:related}
\subsection{Camera-based Physiological measurement}
Since Takano et al. \cite{takano2007heart} and Verkruysse et al. \cite{verkruysse2008remote} evaluated the possibility of measuring HR remotely from facial videos, many researchers have proposed different methods to recover physiological data. Among them, some works consider regions of interest using various techniques, including Blind Source Separation \cite{poh2010non, poh2010advancements, lewandowska2011measuring}, normalized Least Mean Squares \cite{li2014remote} or self-adaptive matrix completion \cite{tulyakov2016self}. In contrast, other works rely on the skin optical reflection model by projecting all RGB skin pixels channels into a more refined subspace mitigating motion artifacts \cite{de2013robust, wang2015novel, wang2016algorithmic}. 

Recently, deep learning-based methods \cite{vspetlik2018visual, yu2019remote, perepelkina2020hearttrack, lee2020meta, lu2021dual, nowara2021benefit} have outperformed conventional methods and achieved state-of-the-art performance estimating vital signs from facial videos. Some of these methods leverage prior knowledge learned from traditional methods and combine it with CNNs to exploit more sophisticated features \cite{niu2018synrhythm, niu2019rhythmnet, song2021pulsegan}. On the other hand, some other researchers have aimed at fully end-to-end approaches \cite{chen2018deepphys, Yu2019RemotePS, perepelkina2020hearttrack}. Unlike previous methods, end-to-end models utilize facial videos as input to directly predict the rPPG signal.

\subsection{Deep Spatio-Temporal modelling}
\label{subsec:temp_modelling}
Spatio-temporal modelling plays a crucial role in rPPG measurement. To learn better PPG features from facial video, DL approaches have been explored under different space-time schemes: sequential (e.g. CNN combined with LSTM \cite{hochreiter1997long ,lee2020meta}) , in parallel (e.g. two-branch CNN \cite{chen2018deepphys, wang2019vision}) or simultaneous (e.g. 3DCNNs \cite{Yu2019RemotePS, perepelkina2020hearttrack}). From these schemes, the most explored one is based on 3DCNNs, which simultaneously capture spatial and temporal information, reaching impressive results in heart rate (HR) estimation. However, they require high computational cost and a significant increase of trainable parameters compared with conventional 2DCNNs. Some alternatives appeared to overcome the computational cost of 3DCNNs while preserving the robustness of Spatio-temporal modelling.
Firstly, Chen et al. \cite{chen2018deepphys} proposed a convolutional attention network (CAN) using normalized frame differences as input to emulate the first-order derivative in the temporal domain. Other studies attempted to boost 3DCNN temporal modelling by incorporating  Temporal Shift Modules (TSM) \cite{liu2020multi, lin2019tsm} and approximating temporal representation employing a combination of 2D and 1D convolutions \cite{liu2020general}. 
In contrast, Liu et al. \cite{yu2020autohr} designed a novel Temporal Difference Convolution (TDC) to simulate the normalized frame difference from \cite{chen2018deepphys}, but at feature level. However, we argue that this approach is not benefiting from higher orders dynamics that can contribute to more refined features from PPG signal recovery. A recent study proposed by Hill et al. \cite{hill2021learning} demonstrated the benefits of incorporating multi-derivative convolutions to obtain better PPG signal estimation. The problem of this method is that it is applying first and second-order derivatives as an input following the architecture presented by \cite{chen2018deepphys}. Unlike these approaches, we propose exploring the contribution of higher-order dynamics into feature level that is currently still largely unexplored.

\subsection{Modelling temporal offsets in Deep Learning based approaches}
\label{subsec:label}
One of the major drawbacks of Deep Learning (DL) approaches for rPPG compared to traditional ones such as \cite{de2013robust,wang2016algorithmic} is the need of a reliable ground truth of the BVP signal during the training process. 
More precisely, the rPPG estimation from facial videos demands a very precise synchronization between facial and physiological recordings. Nevertheless, since most of the existing databases employed finger pulse oximeter to record PPG signals \cite{stricker2014non, bobbia2019unsupervised, zhang2016multimodal, niu2018vipl}, they did not consider the physiological offset or PTT between the facial and finger blood flow \cite{mocco2016skin}.  
Another potential synchronization error that can occur is during the acquisition procedure due to the rolling shutter effect. Mironenko et al. \cite{mironenko2020remote} proved the existence of a slight phase shift between both modalities caused by the progressive scanning of some imaging devices.
Consequently, we are forcing DL models to learn inaccurate correspondences between the spatial and physiological information during the training stage.

Zhan et al. \cite{zhan2020analysis} were the first to determine the impact of phase-shift regarding the HR error between default and phase-corrected reference training labels. Since traditional approaches depend directly on the facial skin pixels, they used a POS approach to correct reference labels through the Hilbert Transform. Despite recent studies \cite{speth2021unifying} are adopting the same strategy with different conventional approaches, this rectification implies an extra preprocessing step relying on the accurate PPG estimation of a traditional method. A different alternative to address the reference offset was presented by Gideon et al. \cite{gideon2021way}, who introduced a frequency loss based on maximum cross-correlation (MCC) to be more robust to temporal offsets. However, they apply the cross-correlation loss at sub-sequence level, which provides too much flexibility to the model by ignoring the fact that temporal offsets are only dependent on the video or subject. In our ablation studies \cref{loss_experiments}, we demonstrate that the benefits of considering temporal offsets at the subject level instead of using a sub-sequence level also imply a better HR estimation.

In this paper, to overcome the phase-shift problem we propose an alternative and simpler loss based directly on the temporal domain, which provides very competitive results compared with the common losses used in rPPG estimation.


%% file: Methodology/Methodology.tex
\section{Methodology}
In this section, we firstly introduce the details of our proposed rPPG end-to-end model. Subsequently, we focus on the presented objective function to deal the temporal offset between facial and physiological data.

\label{sec:method}

\begin{figure*}[t]
    \centering
    \includegraphics[width=160mm,height=50mm]{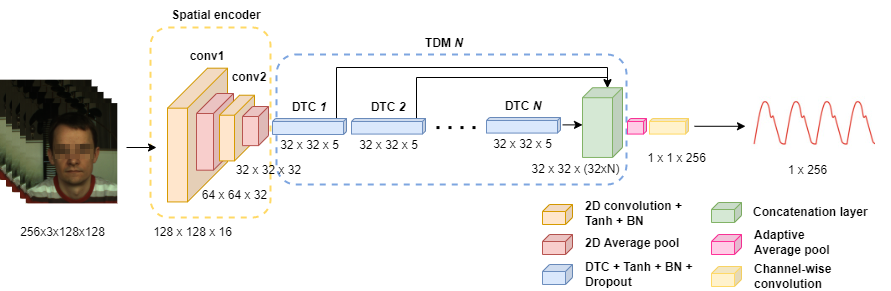}
    \caption{ Overall structure of our proposed model.}
    \label{fig:model}
\end{figure*}

\subsection{Overview}
The motivation of our proposed spatio-temporal model is three-fold: 1) projecting the input RGB images into an optimal latent space encoding the information of the rPPG signal while preserving the model's computational complexity, 2) efficiently modelling temporal information using a cascaded temporal derivative module (TDM), to achieve the equivalent to a high-order Taylor approximation, 3) reduce the influence of PTT in the pulse estimation through TALOS loss. 
Given an input facial RGB video sequence $X$ with size $3 \times T \times H \times W$, the overall architecture is shown in \cref{fig:model}. Here, $3$ refers to RGB channels and $T, H, W$ are the number of frames, height and width of each frame, respectively. 

\subsection{Spatio-temporal Modelling }

The presented network is divided into the spatial encoder and the temporal derivative module. The encoder aims to learn relevant spatial filters that project the facial data into a space relevant to extracting pulsatile information. The encoder consists of two 2D convolutions with a 3x3 kernel size. Each convolution is followed by the hyperbolic tangent activation function, batch normalization (BN) and average pooling operation. Then, the rough spatial features of each frame are reshaped and zero-padded before being fed to the TDM to aggregate the information of all video frames. In our preliminary experiments, we observed that adding more spatial convolutional layers leads to a significantly higher computational complexity during inference without providing a relevant performance improvement. Our proposed TDM is composed of differential temporal convolutions (DTC) that model the dynamics of the spatial features by a chain of derivative filters up to the selected order, with the aim to approximate local dynamics as in a Taylor series expansion. Similar to \cite{hill2021learning}, our architecture employs two consecutive DTC to model first and second temporal order derivatives which represent velocity and acceleration dynamics of rPPG. In order to show the benefits of modelling second-order temporal derivatives, we perform an ablation study where our model is able to extract more sophisticated features using second-order than zero and first order, which are the most common orders in prior work.
To ensure this behavior, we implement the DTC blocks as 1D-convolutions with fixed weights i.e. non-trainable weights chosen to approximate a first-order derivative operator. Therefore, the first order DTC can be formulated as follows:

\begin{equation}
    \mathrm{DTC_1(t) }= \sum_{\tau\in R} \mathrm{w_{\partial t}}(\tau) \cdot \mathrm{x}(t+\tau)
\end{equation}

\noindent where $t$ denotes the current temporal location, $ \mathrm{x}$ is the output tensor from the 2DCNN corresponding to the extracted spatial features,
$ \mathrm{w_{\partial t}} = [-2, -1, 0, 1, 2]$ is the derivative filter using $5\times1$ kernel size and $R$ is the receptive field. Our preliminary experiments revealed that the use of longer temporal windows did not lead to a significant performance improvement. Then, the n-th order DTC is defined recursively as:

\begin{equation}
    \mathrm{DTC_n(t) }= \sum_{\tau\in R} \mathrm{w_{\partial t}}(\tau) \cdot \mathrm{DTC_{n-1}}(t+\tau),\qquad \forall n > 1
\end{equation}

\noindent where the temporal information is modeled using the output tensor of previous DTC instead of the spatial features from the 2D encoder. 
Later, we can define the TDM  as the concatenation across the channel dimension of the tensor of all the DTC derivatives :

\begin{equation}
 \mathrm{TDM = Concat(DTC_1, DTC_2, \ldots, DTC_n)}
\end{equation}

\noindent Finally, the TDM output is projected into signal space using a channel-wise convolution operation with 1×1 kernel to generate the predicted rPPG signal of size $T$.

\label{subsec:der_model}
\subsection{Loss function}
\label{subsec:loss_function}
The impact of PTT during the training process of learning-based methods has been previously discussed in \cite{zhan2020analysis}. In particular, it has been shown that it is crucial  to design an appropriate objective function robust to the temporal offset produced during the acquisition procedure or caused by the PTT between facial and finger regions. For this reason, we introduce the TALOS (\textbf{T}emporal \textbf{A}daptive \textbf{LO}cation \textbf{S}hift) Loss, a new temporal loss to train deep learning models in the context of rPPG signal estimation.

To design our loss function, we consider two assumptions. Firstly, the time offset between facial data and the finger-PPG signal is in the order of milliseconds, and thus it is well bounded within $\pm \frac{F_s}{2}$ frames, where $F_s$ is the video sampling frequency. Secondly, the PPG phase shift is the same for all video instances of the same subject because the time delay depends on each individual's physiology and the recording setting, which remains the same.

Considering the previous assumptions, we propose a loss function designed to be robust against possible unknown temporal offsets. In particular, we introduce a latent variable for each subject $s$ as $K_s = \{k\,|\, \frac{k}{2} \in \mathbb{Z}: -\frac{F_s}{2} \leq k \leq \frac{F_s}{2}\}$, representing all the possible temporal shifts for a given subject. Given the previous definition, our goal is to learn a parametric distribution:
\begin{equation}
p_{\mathbf{\theta}}(k|s) = \frac{\exp(\theta_k^s)}{\sum_{i=1}^K \exp(\theta_k^s)},
\end{equation}
where parameters $\mathbf{\theta}^s \in \mathbb{R}^K$ encode the logits of a multinomial distribution for all the possible temporal offsets $k$ given a subject $s$. Using the previously defined distributions, our proposed TALOS loss function is computed during optimization as described bellow. 

Let $\hat{y}(t)$ be the predicted signal from our model, and $y(t)$ the ground-truth signal, both with the same length $T$. We zero-pad the ground truth signal with respect to all possible offsets $k$:

\begin{equation}
    y_{pad} = \begin{cases}
    pad(k, y) ,& \text{if } k\leq 0 \\ pad(y, k), & \text{if }  k>0
\end{cases}
\end{equation}

Then, the padded ground-truth signal $y_{pad}$ is shifted for all possible delays within $K$ and each $y_k$ is used to compute the Mean Squared Error (MSE) with respect to the predicted signal $y$:

\begin{equation}
    y_k = y_{pad}(t-k) \quad \forall k \in K
\end{equation}

\begin{equation}
    MSE(\hat{y}, {y}_k)=\frac{1}{T} \sum_{t=1}^{T}(\hat{y}(t)-{{y}_k}(t))^2
\end{equation}

Finally, we express our proposed TALOS loss function as:

\begin{equation}
    \mathcal{L}_\mathrm{{TALOS}} = \sum_{k\in K} MSE(\hat{y}, {y}_k) \cdot p_{\mathbf{\theta}}(k|s),
\end{equation}
where the MSE for each possible offset $k$ is weighted according to the learned temporal-shift probability $p_{\mathbf{\theta}}(k|s)$. Note that this probability is dependent on the the subject $s$ corresponding to the specific ground-truth video.

The rational of the above formulation is that, when the value of $k$ is approximately the offset between the ground truth and the pulsatile variation captured by the input video, then the $MSE$ is expected to reach its minimum value. However, $k$ is not known beforehand and thus, we learn a distribution over a latent variable representing all the possible temporal shifts. Note that the parameters $\theta^s$ for this distributions are optimized during training and, as a consequence, we expect to assign a higher probability to offsets $k$ leading to a lower MSE error.   


%% file: Experiments/Experiments.tex
\section{Experiments}
\label{sec:experiments}

In this section, we first introduce the two benchmark datasets used in
the experiments and describe the implementation details. Then, we
compare the impact of different modules explained in \Cref{sec:method} on the test results.
Finally, we present a relative comparison with the existing rPPG methods.
\subsection{Datasets}
\label{subsec:datasets}
We evaluated our approach on the following two RGB video datasets.

\textbf{PURE} \cite{stricker2014non}: The Pulse Rate Detection Dataset contains 60 videos from 10 subjects (eight male, two female) performing six different head motion tasks: steady, talking, slow translation, fast translation, small rotation, and medium rotation. The facial videos were recorded using an ECO274CVGE camera with a resolution of 640 x 480 pixels and 30 frames per second (FPS). Each video is about 1 minute long and stored in uncompressed PNG format. The gold-standard measures of BVP and SpO2 were collected using a finger pulse oximeter (pulox CMS50E) with a sampling rate of 60 Hz. PURE contains predefined splits for training and test (7 subjects for training, 3 subjects for testing), which we use to compare with the current related work. 

\textbf{UBFC-RPPG} \cite{bobbia2019unsupervised}: The UBFC-RPPG dataset includes 42 RGB videos from 42 subjects. The subjects were asked to play a time-sensitive mathematical game, emulating a standard human-computer interaction scenario, to obtain varied HR during the experiment. The recorded facial videos were acquired indoors with varying sunlight and indoor illumination at 30 FPS with a webcam (Logitech C920 HD Pro) at a resolution of 640x480 in uncompressed 8-bit RGB format. The bio-signals ground-truth were acquired using a CMS50E transmissive pulse oximeter to record PPG signal and heart rate with a 60 Hz sampling rate. To validate UBFC, we use the first 28 subjects for training and the last 14 subjects for testing following \cite{lokendra2022and}.

\subsection{Implementation details}
\label{subsec:setup}
\subsubsection{Preprocessing and training procedure}
We adopt the same preprocessing stage for each dataset in all our experiments. Firstly, we estimate facial regions for each frame using the MTCNN \cite{zhang2016joint} algorithm adding a 50\% scale size of the detected box and resizing each frame to $128\times 128$ pixels. The
ground-truth bio-signal is preprocessed following \cite{dall2020prediction} to denoise the raw PPG signal, which facilitates a better model convergence during the training procedure. The resulting bio-signal is then downsampled to 30 Hz so that it matches the sampling rate of the videos.

We implemented our model using Pytorch 1.8.0 \cite{paszke2019pytorch} and trained it on a single NVIDIA GTX1080Ti. We utilized sequences of 256 frames without overlap and Adadelta optimizer during training \cite{zeiler2012adadelta}. In addition, when using our TALOS loss, we incorporated an extra SGD optimizer with a 0.01 learning rate to optimize the parameters $\mathbf{\theta}^s$ of the temporal-shifts distributions for each subject. Finally, the estimated HR was computed from the predicted rPPG signal using the power spectral density (PSD). Before calculating the HR value, we applied a band-pass filter with cutoff frequencies of 0.75Hz and 2.5 Hz, similarly to \cite{chen2018deepphys}.

\subsubsection{Metrics and evaluation}
To evaluate the HR estimation performance of the proposed model, we adopted the same metrics used in the literature, such as
the mean absolute HR error (MAE), the root mean squared
HR error (RMSE) and Pearson’s correlation coefficients R \cite{li2014remote}, \cite{tulyakov2016self}. Besides, we computed the computational performance in terms of the number of parameters and Multiple-Accumulate (MACs) \footnote{\url{https://github.com/sovrasov/flops-counter.pytorch}} to compare the efficiency of our model with respect to the major existing methods.
Ablation experiments were performed using 256-frame sequences (8.5 seconds approx) with no overlap to compute HR estimation and all reported metrics, which is more challenging and informative than HR estimation based on the whole video sequence at once. Nevertheless, we also computed whole-video performance to fairly compare our method to the state-of-the-art, because most prior work adopts whole-video evaluation. Thus, we compare our approach with different traditional and deep learning methods. Since not all learning-based methods report HR results on the same dataset, we re-implemented the PhysNet 
model \cite{Yu2019RemotePS} using the code provided by the author \footnote{\url{https://github.com/ZitongYu/PhysNet}} and the hyper-parameters
in the paper. For DeepPhys \cite{chen2018deepphys} and Zhan et al. \cite{zhan2020analysis} model, the source code is not publicly available, so we also re-implemented them according to the details provided in the respective papers.

\subsection{Ablation studies}

\label{subsec:ablation}
\subsubsection{Evaluation of Spatio-temporal modelling}

Our first experiment to configure our model analyses the influence of aggregating DTCs. For this reason, we evaluate our model into three different schemes. Firstly, we consider our model using only spatial information without modelling temporal dynamics. The 2DCNN spatial encoder forms this first configuration, also called $TDM_0$, since it represents a zero-order derivative. Later, to create high-order models, we introduce sequential DTCs creating $TDM_1$ and $TDM_2$, representing the first and second-order derivatives. \cref{table:derivatives_exp} summarized the sequence HR estimation for each temporal derivative order, which indicates that the aggregation of high-order derivatives is producing a more robust spatio-temporal representation for rPPG.

\begin{table}[!ht]
\renewcommand{\arraystretch}{1.}
\centering
\caption{Evaluation of TDC aggregation}
  \begin{tabular}{c|c c c }
    \hline
    \multirow{2}{1.5cm}{Derivative order} & \multicolumn{3}{c}{Sequence evaluation }  \\ [0.2ex]
    \cline{2-4}
    & MAE$\downarrow$&RMSE$\downarrow$&R$\uparrow$\\
    \hline
    $\mathrm{TDM_0}$ & 4.34 & 8.05 & 0.80 \\ [0.2ex]
    $\mathrm{TDM_1}$ & 2.84 & 6.01 & 0.88 \\ [0.2ex]
    $\mathrm{TDM_2}$  & 2.68  & 5.55 & 0.90 \\[0.2ex]
    \hline
    \end{tabular}
    
    \label{table:derivatives_exp}
    \end{table}

As mentioned in \cref{subsec:temp_modelling}, there are different approaches to model spatio-temporal information. Then, in this second experiment, we fix the first layers of our model (spatial-only), used as a baseline model, and we evaluate different choices for the temporal modelling stage. To this end, we implement two modified versions of our proposed model. In the first version, we adopt the best configuration found in the first experiment, which consists of $TDM_2$, shown in \cref{fig:model}, while in the second version, we combine spatial and temporal information simultaneously, substituting the TDM by two 3D convolutional layers. 
To validate the capability of the different configurations, we evaluate each architecture in terms of sequence HR estimation, but also the performance of each architecture, summarized in \cref{table:spatiotemporal_exp}.   
In terms of HR estimation, our proposed TDM and the modified 3D architecture obtained similar results, while the 2DCNN model produces higher HR error than the other configurations. On the other hand, regarding the computational cost, we observe a significant difference in the number of parameters and MACs between 2DCNN and our TDM approach compared with the 3DCNN configuration. This suggests that constrained temporal modelling based on dynamic properties can capture temporal information as accurately as 3DCNNs but with far less parameters and reduced computational cost, similar to 2DCNN performance.

\begin{table}[!ht]
\caption{Evaluation of spatio-temporal modelling.}
\begin{adjustbox}{width=\columnwidth,center}
\renewcommand{\arraystretch}{1.}
\centering
 
  \begin{tabular}{l|c c c |c c}
    \hline
    \multirow{2}{0.3cm}{Model} & \multicolumn{3}{|c}{Heart Rate} & \multicolumn{2}{|c}{Performance}  \\
    \cline{2-6}
    & MAE$\downarrow$&RMSE$\downarrow$&R$\uparrow$ & Params(K) & MACs(G)\\
    \hline
    2DCNN  &  4.34 & 8.05 & 0.80 & 5.17 & 6.95
 \\ [0.2ex]
    3DCNN  & 2.69 & 5.66 & 0.89 & 15.51 & 9.70
 \\[0.2ex]
    TDM   & 2.68  & 5.55 & 0.90 & 5.26 &  7.08  \\ [0.2ex]
    \hline
    \end{tabular}
    \end{adjustbox}
    \label{table:spatiotemporal_exp}
    \end{table}

\subsubsection{Evaluation of the objective function }
\label{loss_experiments}
To evaluate the effectiveness of TALOS loss, we compare it to other loss functions under the same network configuration explained in the previous section. To this end, we choose the MSE, and Negative Pearson Correlation (NPC) losses \cite{Yu2019RemotePS}, which are widely used in the prior work for remote heart rate estimation.
Furthermore, to assess the robustness of temporal offsets on the ground-truth signal, we also evaluate a frequency domain loss function, the maximum cross-correlation (MCC) loss proposed in \cite{gideon2021way}.

\begin{table}[!hb]

\renewcommand{\arraystretch}{1.}
\centering
\caption{Comparison of different loss functions.}
  \begin{tabular}{c|c c c }
    \hline
    \multirow{2}{0.5cm}{Loss} & \multicolumn{3}{c}{Sequence evaluation }  \\ [0.2ex]
    \cline{2-4}
    & MAE$\downarrow$&RMSE$\downarrow$&R$\uparrow$\\
    \hline
    MSE  & 2.68  & 5.55 & 0.90   \\ [0.2ex]
    NPC \cite{Yu2019RemotePS}  & 2.67  & 5.62 & 0.89 \\[0.2ex]
    MCC \cite{gideon2021way} & 2.44 & 3.73 & 0.96  \\ [0.2ex]
    TALOS  & 2.33 & 3.41 & 0.96 \\ [0.2ex]
    \hline
    \end{tabular}
    
    \label{table:loss_exp}
    \end{table}
    
\begin{figure*}	[!ht]
	\centering
	\begin{subfigure}[t]{2in}
		\centering
		\includegraphics[width=2in]{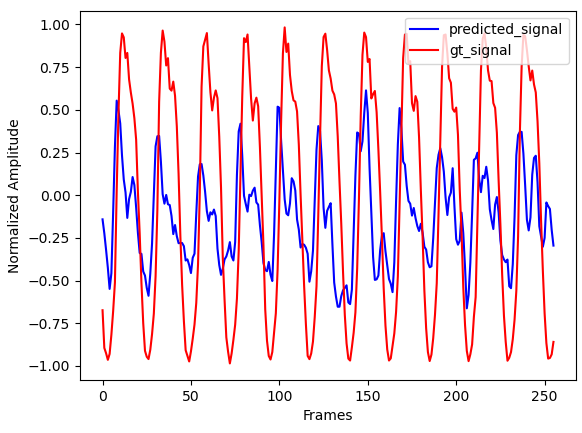}
		\caption{Proposed method trained with $\mathcal{L}_\mathrm{{MSE}}$ loss }\label{fig:reg_mse}
	\end{subfigure}
	\quad
	\begin{subfigure}[t]{2in}
		\centering
		\includegraphics[width=2in]{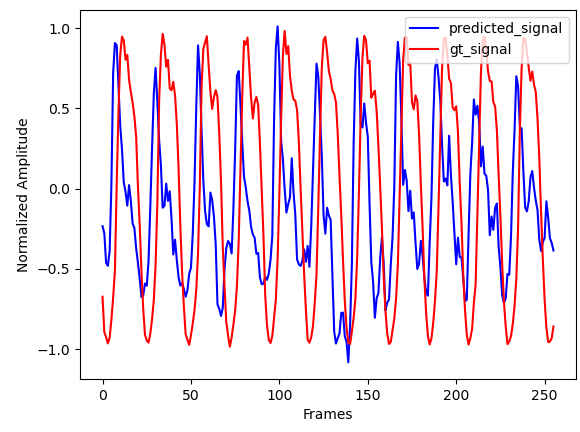}
		\caption{Proposed method trained with $\mathcal{L}_\mathrm{{TALOS}}$ }\label{fig:reg_talos}
	\end{subfigure}
	\begin{subfigure}[t]{2in}
		\centering
		\includegraphics[width=2in]{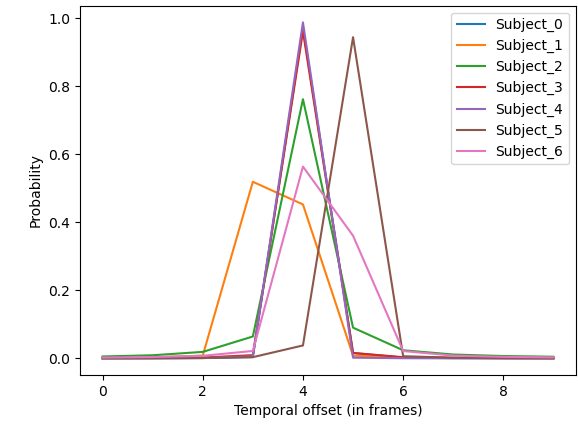}
		\caption{Latent probability matrix $P_{\theta}$}\label{fig:shift_probs}
	    \label{fig:Lp}
	\end{subfigure}
	
	\caption{PPG sequence estimation on individual’s PURE test data using the standard MSE loss, \cref{fig:reg_mse}, and the proposed TALOS loss, \cref{fig:reg_talos}. In \cref{fig:reg_talos} we observe how the model is able to predict the PPG signal in the temporal instance where the regression error is minimum, regardless of the ground-truth phase. On \cref{fig:shift_probs} is shown the learned frame offset for each subject during the training process using TALOS loss. The offset range is limited between 0 and 300 ms for better visualization.}\label{inference}
\end{figure*}

\cref{table:loss_exp} summarizes the results, which show a similar trend between MSE and NPC losses. This can be explained because NPC is only focused on the similarity between signals without considering the morphology and range of the signals, which can lead to some estimation error. Usually, the majority of methods that adopt NPC apply some previous normalization before propagating the loss, which we do not consider to compare all the loss functions in a more direct and fair manner. In contrast, we can appreciate a significant difference between these previous temporal loss functions and temporal offset invariant losses such as TALOS or MCC. These results show the impact on HR estimation of the PTT delay between the ground truth and the facial data, which makes it harder for the model to learn the actual relation between the predicted signal from the skin color changes and the label signal when using conventional loss functions. Finally, comparing MCC with TALOS, we observe that our proposed loss obtains better results in MAE and RMSE. This difference can be explained because MCC considers each sequence or video independently while TALOS loss enforces to learn a consistent shift distribution for each subject. In \cref{inference}, we can note that the predicted signal amplitude, shown in \cref{fig:reg_talos}, is substantially more accurate than the estimates produced under standard losses (\cref{fig:reg_mse}), while the offset between the ground truth and the prediction is still clearly visible. This effect is due the built-in estimate of the shift probabilities of TALOS, which we show in \cref{fig:Lp}: notice how the estimated shift probabilities for each subject always follow a similar pattern dominated by a peak that spreads one or two frames that is surrounded by near-zero values elsewhere. Further, these peaks are localized between $3$ to $5$ frames, suggesting offsets that range from $100$ to $170$ milliseconds, depending on the subject.



\begin{table*}
\renewcommand{\arraystretch}{0.95}
\centering
\caption{: Performance of heart rate measurement for PURE and UBFC-rPPG datasets.}
  \begin{tabular}{c|c c c |c c c }
    \hline
    \multirow{2}{1cm}{Method} & \multicolumn{3}{|c}{PURE } & \multicolumn{3}{|c}{UBFC}\\ 
    \cline{2-7}
    & MAE$\downarrow$&RMSE$\downarrow$&R$\uparrow$  
    & MAE$\downarrow$&RMSE$\downarrow$&R$\uparrow$ 
    \\ 
    \hline
    
    CHROM \cite{de2013robust}  & 2.07 & 2.50 & 0.99 & 3.44 & 4.61 & 0.97  \\[0.2ex]
    POS \cite{wang2016algorithmic}  & 3.14 & 10.57 & 0.95 & 2.44 & 6.61 & 0.94 \\[0.2ex]
    HR-CNN \cite{vspetlik2018visual}   & 1.84 & 2.37 & 0.98 & - & - & -    \\ [0.2ex]
    DeepPhys \cite{chen2018deepphys}   & 1.84 & 2.31 & 0.99 & 2.90 & 3.63 & 0.98   \\ [0.2ex]
    PhysNet  \cite{Yu2019RemotePS}  &   2.16 & 2.70 & 0.99 & 2.95 & 3.67 & 0.98    \\ [0.2ex]
    Zhan et al. \cite{zhan2020analysis}   & 1.82 & 2.29 & 0.99 & 2.44  & 3.17 & 0.99   \\ [0.2ex]
    Gideon et al. \cite{gideon2021way}  & 2.30 & 2.90 & 0.99 & - & - & -    \\ [0.2ex]
    AND-rPPG\cite{lokendra2022and} & - & - & - & 2.67 & 4.07 & 0.96  \\ [0.2ex]
    \hline
    \textbf{Ours} & 1.89 & 2.33 & 0.99 &  2.54 & 3.31 & 0.99    \\ [0.2ex]
    \textbf{Ours + TALOS} &  1.83 & 2.30 & 0.99  & 2.32 & 3.08 & 0.99    \\ [0.2ex]
    \hline
    \end{tabular}
    
    \label{table:sota}
    \end{table*}

\subsection{Comparison with existing methods}
\label{subsec:sota}

After the preliminary studies from the previous sections, we select our best model configuration to compare against other reported results on the PURE and UBFC-rPPG datasets. Besides, we compare the efficiency of our model with respect to the reproduced methods explained on \cref{subsec:setup}. In contrast to the ablation studies of \cref{subsec:ablation}, we estimate the average HR for each video as a whole, to allow for fair comparison to results from other researchers. Results are summarized in \cref{table:sota}.

\textbf{1) Results on PURE dataset:} We first conduct the video HR evaluation on the PURE dataset. These results show that DL methods like HR-CNN and DeepPhys perform much better than the hand-crafted methods such as POS and CHROM. This suggests that learning-based models can extract more representative features than hand-crafted methods for PPG signal and HR estimation. Indeed, the results of the compared DL approaches vary between 2.29 and 2.90 BPM RMSE, which denotes the good performance in non-compressed data and controlled scenarios such as the PURE dataset. Regarding our proposed method, we can appreciate that it outperforms some DL methods such as HR-CNN or PhysNet, achieving at the same time similar results as Zhan et al. \cite{zhan2020analysis} and DeepPhys, which obtains the best results. 

\textbf{2) Results on UBFC-rPPG dataset:} We also evaluate our model in the public UBFC dataset. Comparing these results to those from the PURE dataset, we observe that UBFC is more challenging since the RMSE from the different methods varies between 3.08 and 6.61 BPM. It can be seen that our proposed model achieves the best HR video results concerning both traditional and DL approaches. This performance is possible through TALOS loss, which considerably improves our baseline results by adopting a standard MSE regression. Comparing the impact of TALOS loss in HR video estimation for PURE and UBFC datasets, we note a significant error decrease in UBFC and a more modest improvement in the PURE dataset. The limited number of subjects can explain this difference in the PURE dataset, which consists of several recordings from only ten subjects, while UBFC datasets contain more subject diversity and fewer recordings per user.    

\textbf{3) Results on computation cost:} Finally, we analyze the performance efficiency of our method compared with the publicly available and re-implemented DL methods. \cref{table:performance} shows that our lightweight model is extremely efficient with respect to the state-of-the-art models in terms of parameters and MACs. Considering this reduced computational cost, our model outperforms the PhysNet, consisting of a 3DCNN Spatio-temporal auto-encoder. This means that a reduced model with well-designed but constrained temporal modelling, and suitable PPG preprocessing can achieve similar or better results than 3DCNNs, which usually require massive data to train all the model parameters.

\begin{table}[!ht]
\centering
\renewcommand{\arraystretch}{1.1}
\caption{Efficiency performance on tested methods.}
\begin{tabular}{c cc}
\hline
Methods & Params(K) & MACs(G)  \\
\hline
DeepPhys \cite{chen2018deepphys}  &  1060  & 16.44  \\ 
PhysNet  \cite{Yu2019RemotePS} & 768 & 112.33 \\ 
Zhan et al. \cite{zhan2020analysis} & 2620  & 59.40  \\ 
\textbf{Ours} & 5.26 &  7.080 \\
\hline  
\end{tabular}

\label{table:performance}
\end{table}

%% file: Conclusions/Conclusions.tex
\section{Conclusions}
\label{sec:conclusions}

In this paper, we propose a lightweight model for remote heart rate measurement. The presented method effectively estimates rPPG spatio-temporal features by aggregating multiple temporal derivative filters up to the desired order. Moreover, to mitigate the effect of possible temporal offsets such as PTT, we introduce a new objective loss function, which we call TALOS, designed to learn the latent shift probabilities of the ground truth during the training process, which allows to minimize the error between the predicted and label signals under the estimated relative offset. Through experiments on the PURE and UBFC-rPPG datasets, our proposed framework demonstrates a competitive HR performance with reduced computational requirements, which facilitates applicability to real/low-resource scenarios. In our future work, we aim to explore the combination of higher-order dynamics to further improve the temporal modelling for rPPG estimation. 

%% file: Acknowledgment/Acknowledgment.tex
\section*{Acknowledgments}
This work is partly supported by the eSCANFace project (PID2020-114083GB-I00) funded by the Spanish Ministry of Science and Innovation.